\documentclass[twocolumn]{autart}    

  \let\theoremstyle\relax
  \usepackage{soul} 
\usepackage{amsmath}
\usepackage{mathtools}
\usepackage{amsthm}
\usepackage{amssymb}
\usepackage{accents}
\usepackage{cite}

\usepackage{multirow}
\usepackage[bookmarks=true]{hyperref}
\hypersetup{%
  breaklinks,
  bookmarksnumbered=true,
  bookmarksopen=true,
  colorlinks=true,
  linkcolor=black,
  urlcolor=black,
  citecolor=black
}
\usepackage{siunitx}
\usepackage{graphicx}
\usepackage{algorithm}
\usepackage{algpseudocode}
\newcommand\numberthis{\addtocounter{equation}{1}\tag{\theequation}}

\newcounter{theorem}
\newcounter{definitioncount}

\newtheorem{proposition}[theorem]{Proposition}

\newtheorem{definition}[definitioncount]{Definition}

\begin{document}

\begin{frontmatter}
\title{The Salted Kalman Filter:\\
Kalman Filtering on Hybrid Dynamical Systems\thanksref{footnoteinfo}}

\thanks[footnoteinfo]{This material is based upon work supported by the U.S. Army Research Office under grant \#W911NF-19-1-0080 and the National Science Foundation under grants \#IIS-1704256 and \#ECCS-1924723. The views and conclusions contained in this document are those of the authors and should not be interpreted as representing the official policies, either expressed or implied, of the Army Research Office, National Science Foundation, or the U.S. Government. The U.S. Government is authorized to reproduce and distribute reprints for Government purposes notwithstanding any copyright notation herein. The material in this paper was not presented at any conference. Corresponding author N.~J.~Kong.}

\author[cmu]{Nathan J. Kong}\ead{njkong@andrew.cmu.edu},    
\author[cmu]{J. Joe Payne}\ead{jjpayne@andrew.cmu.edu},               
\author[cmu]{George Council}\ead{gcouncil@andrew.cmu.edu},  
\author[cmu]{Aaron M. Johnson}\ead{amj1@cmu.edu}  

\address[cmu]{Department of Mechanical Engineering, Carnegie Mellon University, Pittsburgh, Pennsylvania}  


\begin{abstract}                          
Many state estimation and control algorithms require knowledge of how probability distributions propagate through dynamical systems.
However, despite hybrid dynamical systems becoming increasingly important in many fields, there has been little work on utilizing the knowledge of how probability distributions map through hybrid transitions.
Here, we make use of a propagation law that employs the saltation matrix (a first-order update to the sensitivity equation) to create the Salted Kalman Filter (SKF), a natural extension of the Kalman Filter and Extended Kalman Filter to hybrid dynamical systems.
Away from hybrid events, the SKF is a standard Kalman filter. 
When a hybrid event occurs, the saltation matrix plays an analogous role as that of the system dynamics, subsequently inducing a discrete modification to both the prediction and update steps.
The SKF outperforms a naive variational update -- the Jacobian of the reset map -- by having a reduced mean squared error in state estimation, especially immediately after a hybrid transition event.
Compared a hybrid particle filter, the particle filter outperforms the SKF in mean squared error only when a large number of particles are used, likely due to a more accurate accounting of the split distribution near a hybrid transition.
\end{abstract}

\end{frontmatter}

\section{Introduction}
    \begin{figure}[tb]
    \centering
    \includegraphics[width = \columnwidth]{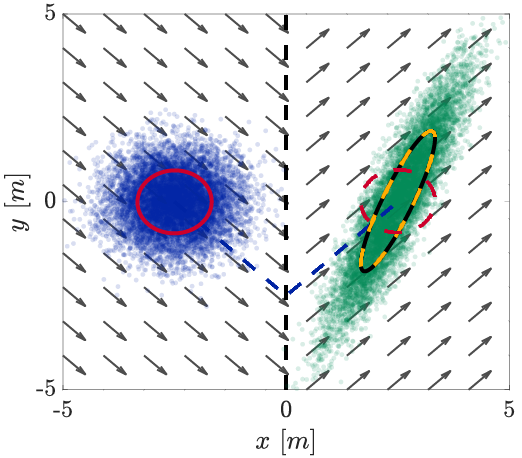}
    \caption{Flowing an initial distribution (blue dots) with covariance (red solid line) along a nominal trajectory (blue dashed line) through hybrid systems with dynamics (arrows) and a single guard (black dashed line). The final distribution (green dots) is overlaid with the actual covariance (black line). Estimated covariance using the Jacobian of the reset map (red dashed line) is compared against our proposed estimate using the saltation matrix (gold dashed line).}
    \label{fig:covariance_mappings}
\end{figure}

From legged robots to manipulator systems, many important contemporary control problems revolve around systems that make and break contact with their environments. 
These contact events are often represented as a discrete change to the system dynamics which introduces complexity for state estimation and control, as classic methods assume smoothness \cite{bloesch2013state,hartley2020contact,mitcheetahstateestimation2018,varin2018constrained}. 
These ``hybrid systems'' \cite{Back_Guckenheimer_Myers_1993,LygerosJohansson2003,goebel2009hybrid} are systems with both continuous states (such as the position and velocity of a robot's center of mass and joints) and discrete states (such as whether or not a limb is in contact with the ground).
Lacking out-of-the-box solutions, state estimation for these  systems is a frontier with novel difficulties \cite{blom1988interacting,skaff2005contextbased}, including how to deal with nonlinear dynamics on the continuous phases \cite{barhoumi2012observer}, discrete jumps in the continuous state \cite {balluchi2013design}, and real time computation \cite{ZHANG2020}.

In this work we propose a Kalman-like filter compatible with hybrid dynamical systems
while also avoiding the combinatorial effects of considering multiple modes simultaneously \cite{ZHANG2020}.
To do this, we apply the saltation matrix (a standard tool from non-smooth analysis \cite{leine2013dynamics}) to propagate state uncertainty covariance through hybrid transitions \cite{biggio2014accurate}. 
The saltation matrix provides a first order approximation of the effects of a hybrid domain change based on the dynamics in the individual modes, the reset functions, and the location of the reset. 
It might be assumed that the propagation of uncertainty through hybrid transitions could be approximated by simply examining the first order approximation of the reset map itself, i.e.\ the Jacobian of the reset map.
For example, \cite{hartley2020contact} and \cite{bloesch2013state} assume that the hybrid transition does not affect the second moment of the distribution; i.e the reset map is identity and therefore the Jacobian would be an identity matrix.
However, this approach does not take into account the differing dynamics in the distinct modes.
The inaccuracy of the naive approach can be seen in Fig. \ref{fig:covariance_mappings}, where the system has an identity reset map but keeping the second moment constant through the transition does not capture the effect of the hybrid transition on the distribution. 
As such, attempting to use the Jacobian of the reset map, while a ``natural'' idea, is ultimately incorrect. 

The remainder of this paper is organized in the following manner. 
Section \ref{section:RelatedWork} provides a brief review of the hybrid system estimation literature. 
Section \ref{section:ProblemFormulation} defines the problem that we seek to solve in this work as well as establishing the notation and conventions used. 
Section \ref{section:KalmanFiltering} introduces the ``Salted Kalman Filter'' (SKF), which is a Kalman Filter augmented with the capability to propagate the estimated first and second moments through hybrid transitions. 
Section \ref{section:Experiments} explains the experiments used to validate the performance of the Kalman filter.
Section \ref{section:Results} compares results from using the SKF to results using the Jacobian of the reset map and to a particle filter.
Finally, Section \ref{section:Conclusion} provides a discussion of the work presented and potential future work.

\section{Related Work}
\label{section:RelatedWork}
There has been a variety of work on the topic of state estimation for systems with differing dynamics and discrete modes, however current approaches either do not consider systems with state-driven mode transitions (i.e.\ are limited to the ``switched system'' case) \cite{blom1988interacting,blom2004particle,balluchi2002design,skaff2005contextbased,eras2019equality,hwang2006state} or are computationally expensive and difficult to run in an online filtering setting \cite{koval2017manifold,ZHANG2020}. 

Our work seeks to understand how distributions are propagated through state-driven hybrid dynamical systems by applying knowledge from non-smooth systems literature \cite{johnson2016hybrid,aizerman1958determination,hiskens2000trajectory} in order to make simplifying assumptions which retain sufficient information for the purposes of online state estimation. 

\subsection{Hybrid System Estimators}
One approach to filtering on hybrid systems with linear dynamics is to use a filter bank where a filter is assigned to each discrete mode and the output of the filter with the lowest residual is used as the current state estimate \cite{balluchi2002design}. 
Another style of filter bank method mixes the outputs of individual filters by utilizing a probability weight calculated based on measurement residuals and \emph{a posteriori} estimate likelihoods such as the interacting multiple model (IMM) \cite{blom1988interacting}. 
These filtering methods have been extended to hybrid systems with nonlinear dynamics \cite{barhoumi2012observer} and hybrid systems with non-identity reset maps during hybrid transitions \cite{balluchi2013design}. 
However, these filtering bank strategies consider hybrid systems with transitions that do not depend on continuous state and therefore do not account for the effect that the continuous state dependent transitions have on the distribution.
This is an issue because the first 2 moments of the distribution are not guaranteed to be captured after a transition.

Particle filtering approaches seek to represent uncertainty distributions directly with a variety of sample points rather than by representing belief as a parametric (e.g.\ Gaussian) distribution \cite{koutsoukos2002monitoring,koval2017manifold}.
One of the major drawbacks of particle filters and related methods is that they are computationally expensive - may require ($\mathcal{O}(2^n)$) where $n$ is the number of states \cite{thrun2002probabilistic}. Because of this, it may be difficult to utilize them in a real-time setting.

Some optimization based methods seek to circumvent this issue of computational complexity by simultaneously selecting the continuous and discrete states over all timesteps to minimize the error associated with the measurements and the dynamics\cite{ZHANG2020,ferrari2002moving}.
The resulting optimization problem requires a much higher computational load compared to causal forward time stepping methods such as Kalman filters and finite impulse response filters whose computational burden is polynomial in the dimension of the state, e.g.\ methods that rely on a fixed finite number of matrix products per timestep and as such may be limited to offline estimation settings.

Online state estimation methods have been created for complex systems with continuous states and discrete modes, such as the case for legged robots making and breaking contact with the ground \cite{hartley2020contact,bloesch2013state}. 
In these settings, an extended Kalman filter is used to estimate the continuous states and the discrete mode is directly measured through contact sensors. 
The primary focus of these works is on the continuous phases rather than the discrete mode transitions due to the presence of direct mode sensing.
Therefore, these estimators do not directly work for general hybrid systems, because there might not be a sensor to determine the hybrid event and there might be discontinuous jumps in the state. 

\subsection{Non-smooth systems and the saltation matrix}
This work makes extensive use of the saltation matrix \cite{aizerman1958determination,hiskens2000trajectory,leine2013dynamics,burden2018contraction}, which is a discontinuous update to the variational equation solution \cite{khalil2002nonlinear} and is a key part of linearizing hybrid dynamics around a chosen trajectory.
They have previously been used to analyze stability of periodic solutions \cite{aizerman1958determination}, trajectory sensitivity \cite{hiskens2000trajectory}, and infinitesimal contraction \cite{burden2018contraction}. 
Most importantly for this work, the saltation matrix has also been used to derive a covariance propagation update law for mapping distributions through hybrid transitions \cite{biggio2014accurate}.

\section{Problem Formulation}
\label{section:ProblemFormulation}
The specific problem we seek to address in this work is the estimation of continuous states of a hybrid dynamical system given:
\begin{enumerate}
    \item A model of the dynamics in each mode.
    \item A model of how the state resets between modes.
    \item The location of the hybrid guards.
    \item Measurements of the system's continuous state.
\end{enumerate} 
We are specifically not considering:
\begin{enumerate}
    \item The probability of the discrete state.
    \item Hybrid systems with intersecting guards \cite[\S~3-4]{scholtes2012introduction} (e.g.\ in a walking system when multiple feet impact simultaneously).
\end{enumerate} 
As many of these terms have multiple possible mathematical meanings, in this section we provide the essential definitions used in this work.

While there are many similar definitions for a hybrid dynamical system, e.g.~\cite{Back_Guckenheimer_Myers_1993,LygerosJohansson2003,goebel2009hybrid},
in this work we define a \emph{$C^r$ hybrid dynamical system}, closely following \cite[Def.~2]{johnson2016hybrid}:
\begin{definition} \label{def:hs}
    A $C^r$ \textbf{hybrid dynamical system}, for continuity class $r\in \mathbb{N}_{>0} \cup \{\infty,\omega \}$, is a tuple $\mathcal{H} := (\mathcal{J},{\mathnormal{\Gamma}},\mathcal{D},\mathcal{F},\mathcal{G},\mathcal{R})$ where the constituent parts are defined as:
    \begin{enumerate}
        \item $\mathcal{J} := \{I,J,...,K\} \subset \mathbb{N}$ is the finite set of discrete \textbf{modes}.
        \item $\mathnormal{\Gamma} \subset \mathcal{J}\times\mathcal{J}$ is the set of discrete \textbf{transitions} forming a directed graph structure over $\mathcal{J}$.
        \item $\mathcal{D}:=\amalg_{{I}\in\mathcal{J}}$ ${D}_{I}$ is the collection of \textbf{domains} where $D_I$ is a $C^r$ manifold with corners \cite{Joyce2012,Lee2012}.
        \item $\mathcal{F}:= \amalg_{I\in\mathcal{J}} F_I$ is a collection of $C^r$ time-varying \textbf{vector fields}, $F_I:  \mathbb{R}\times D_I\to\mathcal{T}D_I$.
        \item $\mathcal{G}:=\amalg_{(I,J)\in\mathnormal{\Gamma}}$ $G_{(I,J)}(t)$ is the collection of \textbf{guards}, where $G_{(I,J)}(t)\subset D_I$ for each $(I,J)\in \mathnormal{\Gamma}$ is defined as a sublevel set of a $C^r$ function, i.e.\ $G_{(I,J)}(t)= \{x \in D_I|g_{(I,J)}(t,x)\leq0\}$.
        \item $\mathcal{R}:\mathbb{R}\times \mathcal{G}\rightarrow \mathcal{D}$ is a $C^r$ map called the \textbf{reset} that restricts as $R_{(I,J)}:=\mathcal{R}|_{G_{(I,J)(t)}}:G_{(I,J)}(t)\rightarrow D_J$ for each $(I,J)\in \mathnormal{\Gamma}$.
    \end{enumerate}
    \end{definition}
    An execution of a hybrid system \cite[Def.~4]{johnson2016hybrid} starts with initializing a state in some hybrid domain $\mathcal{D}_I$, where $I$ is a discrete mode in $\mathcal{J}$. The dynamics on $I$, $F_I$, are followed until the trajectory reaches a guard $G_{(I,J)}$, where $(I,J)$ is a discrete transition in $\mathnormal{\Gamma}$. 
    This triggers the hybrid transition from mode $I$ to mode $J$ and the reset map $R_{(I,J)}$ is applied to the state to initialize the new state in hybrid domain $\mathcal{D}_J$. The execution is defined over a hybrid time domain \cite[Def.~3]{johnson2016hybrid}, which is a disjoint union of closed time intervals where the start and end of an interval is labeled with an under or over bar $[\underbar{t}_{i},\bar{t}_{i}]$.

    A classic result \cite[Thm.~1,\S~15.2]{hirsch2012differential} for a smooth system $\dot{x} = f(x)$ is that we can linearize around a  trajectory $\phi^t(x)$ using the so-called \emph{variational equation}
    \begin{equation}
        \frac{d}{dt} D_x \phi^t(x_0) = D_xf (\phi^t(x_0)) D_x \phi^t (x_0)
    \end{equation}
    where $D_x$ is the Jacobian with respect to $x$. 
    
    For the type of hybrid systems we consider, an analogous equation exists, but additional care must be taken to treat hybrid events consistently.   
    As shown in \cite{aizerman1958determination, bernardo2008piecewise, hiskens2000trajectory, leine2013dynamics}, if for some time $\tau$ the execution $\phi^\tau(x_0)$ intersects a single surface of discontinuity $G_{(I,j)}$ at time $\bar{t}_i$, the variational equation must be updated discontinuously with the so-called \emph{saltation matrix} 
    $\Xi_{(I,J)}(\bar{t}_{i},x(\bar{t}_{i}))$, which is defined at time $\bar{t}_i$ such that state $x(\bar{t}_{i}) \in G_{(I,J)}$.
    \begin{definition}[{\cite[Prop. 2]{burden2018contraction}}]
    The \textbf{saltation matrix},
    \begin{equation}
        \Xi := D_x R+\frac{\left(F_J-D_xR\cdot F_I - D_tR\right)  D_x g}{D_t g +D_x g \cdot F_I} \label{eq:saltationmatrix}
    \end{equation}
    where 
    \begin{align*}
        \Xi &:= \Xi_{(I,J)}(\bar{t}_{i},x(\bar{t}_{i})), \qquad \enspace F_I := F_I(\bar{t}_{i},x(\bar{t}_{i}))\\
        D_x R &:= D_x R_{(I,J)}(\bar{t}_i,x(\bar{t}_{i})), 
        \: D_t R := D_t R_{(I,J)}(\bar{t}_i,x(\bar{t}_{i}))\\ 
        D_xg &:= D_xg_{(I,J)}(\bar{t}_i,x(\bar{t}_{i})),
        \enspace \: D_tg := D_tg_{(I,J)}(\bar{t}_i,x(\bar{t}_{i}))\\
        F_J &:= F_J(\underbar{t}_{i+1},{R}_{(I,J)}(\bar{t}_i,x(\bar{t}_{i}))
    \end{align*}
    is the first order approximation of variations at hybrid transitions from mode $I$ to $J$ and maps perturbations to first order from pre-transition $\delta x(\bar{t}_{i})$ to post-transition $\delta x(\underbar{t}_{i+1})$ during the $i$th transition in the following way\footnote{For a detailed description of the saltation matrix and its role in linearization, see \cite{leine2013dynamics}.},
    \begin{equation}
        \delta x(\underbar{t}_{i+1}) = \Xi_{(I,J)}\big(\bar{t}_{i},x(\bar{t}_{i})\big) \delta x(\bar{t}_{i}) + \text{h.o.t.}
        \label{eq::saltperturbation}
    \end{equation}
    where \emph{h.o.t.}\ represents higher order terms, i.e, $o(||\delta x||)$.
    \end{definition}

Hybrid systems of the type given in Def. \ref{def:hs} can exhibit complex behavior including sliding \cite{jeffrey2014dynamics}, branching \cite{simic2000towards}, Zeno, and more. 
To ensure that the saltation matrix is well defined for all transitions, we accept the assumptions (which are conventional, e.g., \cite{aizerman1958determination, bernardo2008piecewise, leine2013dynamics, burden2016event}) enumerated  in \cite[Assumptions. 1]{burden2018contraction} to limit the class of hybrid dynamic systems under consideration to possess piecewise-smooth trajectories.
In particular, a key assumption is that transitions are \emph{transverse},~i.e., 
\begin{align}
 &\frac{d}{dt} g_{(I,J)}(t,x(t))=\nonumber\\ 
 &\qquad D_tg_{(I,J)}(t,x)+D_x g_{(I,J)}(t,x)\cdot F_{I}(t,x) < 0,\label{asm:salt-exists}
\end{align}
 
 
Note that \eqref{asm:salt-exists} restricts the definition of the guard from Def.~\ref{def:hs} to be both a sublevel set and only exist when the vector field is transverse to it at the boundary. 
That is, we can write each guard set $G_{(I,J)}$ as the following, where $g:=g_{(I,J)}$, and $x(t)$ is a trajectory in $D_I$
\begin{equation}\label{eq:fullguard}
        G_{(I,J)}:= \left\{x\in D_I ~\middle| ~g(t,x)\leq0, 
        \frac{d}{dt} g(t,x(t)) < 0\right\} 
\end{equation}
Intuitively, transversality implies that trajectories initialized nearby a given $G_{(I,J)}$ undergo exactly one transition for small times.
This assumption also ensures the denominator in \eqref{eq:saltationmatrix} does not approach zero.
    
With these definitions and assumptions, we can now apply the saltation matrix to propagate covariance \cite[Eq. 17]{biggio2014accurate} as part of a dynamic update of a probability distribution at a hybrid transition,
\begin{proposition}\label{proposition:covprop}
When the higher order terms are zero, the mean $\mu$ and covariance $\Sigma$ of a hybrid system at the time of a reset are updated as,
\begin{align}
    \mu(\underbar{t}_{i+1}) &= R_{(I,J)}(\bar{t}_{i},\mu^*)
    \label{eq::resetdynamicupdate}\\
    \Sigma(\underbar{t}_{i+1}) &= 
    \Xi_{(I,J)}(\bar{t}_{i},\mu^*)  \Sigma(\bar{t}_{i},\mu^*)  \Xi_{(I,J)}(\bar{t}_{i},\mu^*)^T
    \label{eq::covdynamicupdate}
    \end{align}
    where $\mu^* :=\mu(\bar{t}_{i})$.
 \end{proposition}
\section{Kalman filtering for hybrid systems}
\label{section:KalmanFiltering}
In this section, we present the Salted Kalman Filter (SKF) by applying Prop.~\ref{proposition:covprop} on the mapping of second moments to Kalman filters, which enables their use on hybrid dynamical systems.
First, we assume $\forall I, ~ F_I(t,x) = A_I(t)x+B_I(t)u(t)$, i.e.\ each mode's vector field is linear. 
Note that for a non-linear or linear time varying $F_I$, $A_I$ and $B_I$ are obtained through sampling. Discretized linear matrices with timestep $\Delta$ are denoted with $A_{I,\Delta}$ and $B_{I,\Delta}$.
To simplify expressions for discrete timesteps, we abuse notation and use $a(k) := a(t_k)$ for any relevant function $a$.
Without loss of generality, we assume the case $u(k) = 0 \:\forall\: k$.
To start, the stochastic difference equations considered for the standard Kalman filter \cite[Eqn.~1.1]{welch1995introduction} on domain $I$ for a hybrid dynamical system with linear dynamics are given by
\begin{equation}
    {x}(k+1) := A_{I,\Delta}{x}(k) + \omega_{I,\Delta}(k)
    \label{eq::cont_dynamics}
\end{equation}
where the process noise, $\omega_{I,\Delta}$, is sampled from a zero mean Gaussian distribution with covariance $W_{I,\Delta}$ at each timestep where the effect of the noise is constant throughout the timestep and is handled by integration.
\begin{align}
    f_{I,\Delta}(x,u,\omega(k)) = \int_{t_k}^{t_{k}+\Delta}\left(F_{I}(t,x,u) + \omega(k)\right)dt\label{eq:noiseintegration}
\end{align}
The stochastic measurement equation \cite[Eqn.~1.2]{welch1995introduction} is defined to be
\begin{equation}
    {y}(k) := C_{I}{x}(k) + v_I(k)
    \label{eq::measurement}
\end{equation}
where $C_I$ is the measurement matrix, and $v_I$ is the measurement noise that is  sampled from a zero mean Gaussian distribution with covariance $V_I$. 

The standard Kalman filter consists of two parts:
the \emph{a priori} update,
\begin{align}
    \hat{x}(k+1|k) &= A_{I,\Delta}\hat{x}(k) \label{eq:apriorix}\\
    \hat{\Sigma}(k+1|k) &= A_{I,\Delta}\hat{\Sigma}(k)A_{I,\Delta}^T + W_{I,\Delta} \label{eq:apriorisigma}
\end{align}
and the \emph{a posteriori} update,
\begin{align}
    &K_{k+1} = \hat{\Sigma}(k+1|k)C_I^T\left[C_I\hat{\Sigma}(k+1|k)C_I^T + V_I\right]^{-1}
    \label{eq::kalmangain}\\
    &\hat{x}(k+1|k+1) = \hat{x}(k+1|k) \label{eq::postmean}\\ &\qquad \qquad \qquad \quad+ K_{k+1}\left[y(k+1) - C_I\hat{x}(k+1|k)\right] \nonumber\\
    &\hat{\Sigma}(k+1|k+1) = \hat{\Sigma}(k+1|k) - K_{k+1}C_I\hat{\Sigma}(k+1|k)
    \label{eq::postcov}
\end{align}
where $K_{k+1}$ is the Kalman gain  \cite[Eqns.~1.9--1.13]{welch1995introduction}.

While the standard Kalman filter is adequate when a trajectory is confined to a single domain, we must also account for hybrid events. In this setting, we assume that the true time of impact to the guard $\bar{t}_{i}$ is unknown to the filter and is estimated by determining when a hybrid transition occurs for the mean. 
In this filter, we allow both the \emph{a priori} and \emph{a posteriori} update to trigger a hybrid transition. Therefore, both updates are modified such that the mean and covariance are properly transformed during the hybrid transition. 

In this section we first show these changes for a Kalman filter on a hybrid dynamical system with linear dynamics (Sec.~\ref{section:apriori}--\ref{section:aposteriori}), then the same changes are similarly applied for the Extended Kalman filter on general hybrid dynamical systems (Sec.~\ref{section:ekf}).

\subsection{Hybrid transition during \emph{a priori} update}
\label{section:apriori}
For the \emph{a priori} update, the state is propagated from the previous estimate for a single timestep $\Delta$. 
If the guard and transversality conditions \eqref{eq:fullguard} are not met during the propagation, no hybrid transition is considered and the standard update is used \eqref{eq:apriorix}--\eqref{eq:apriorisigma}. 
If the conditions are met for the estimated mean trajectory, then the forward simulation is stopped and the time of impact $\bar{t}_{i} = t_k + \Delta_1$ is estimated to be the stopping time -- where $\Delta_1=\bar{t}_{i} - t_k$ and $\Delta_2=t_{k+1}-\bar{t}_{i}$ denote the sub-timesteps such that $\Delta_1+\Delta_2=\Delta$. 
Because we assume that a finite number of isolated transitions occur, this process can be repeated until the entire timestep is simulated. 
Without loss of generality, in this section we only consider the case where a single transition occurs, but appending additional transitions can be computed in a similar fashion.

If a  transition occurs from mode $I$ to mode $J$, the stochastic dynamics \eqref{eq::cont_dynamics} are defined to be,
\begin{align*}
    {x}(k+1) := &A_{J,\Delta_2}\left(R_{(I,J)}\left[A_{I,\Delta_1}{x}(k)+ \omega_{I,\Delta_1}(k)\right]\right. \\
    &\left.+ \omega_{R_{(I,J)}}(k)\right) + \omega_{J,\Delta_2}(k)
    \numberthis
    \label{eq::reset_dynamics}
\end{align*}
where $\omega_{R_{(I,J)}}$ is the reset process noise, sampled from a zero mean Gaussian distribution with covariance $W_{R_{(I,J)}}$, $\omega_{I,\Delta_1}$ is the process noise in domain $I$ with timestep $\Delta_1$, and $\omega_{J,\Delta_2}$ is the process noise in domain $J$ with timestep $\Delta_2$. 
The dynamic update at transition \eqref{eq::resetdynamicupdate}--\eqref{eq::covdynamicupdate} augmented with the reset process noise is, 
\begin{align}
    x(\underbar{t}_{i+1}) =& R_{(I,J)}x(\bar{t}_{i})
    \label{eq::kfresetdynamicupdate}\\
    \Sigma(\underbar{t}_{i+1}) =&\Xi_{(I,J)}\Sigma(\bar{t}_{i})\Xi_{(I,J)}^T +W_{R_{(I,J)}}
    \label{eq::kfcovdynamicupdate}
\end{align}
where the saltation matrix is evaluated at $\Xi_{(I,J)}=\Xi_{(I,J)}(\bar{t}_{i},x(\bar{t}_{i}))$.
Combined with the continuous \emph{a priori} updates before and after transition, \eqref{eq:apriorix}--\eqref{eq:apriorisigma}, the \emph{a priori} update over a full timestep is,
\begin{align}
    \hat{x}(k+1|k) =& A_{J,\Delta_2}R_{(I,J)}A_{I,\Delta_1}\hat{x}(k)
    \label{eq::kf_mean_apriori}\\
    \hat{\Sigma}(k+1|k) =&  A_{J,\Delta_2}[\Xi_{(I,J)}(A_{I,\Delta_1}\Sigma(k)A_{I,\Delta_1}^T
    \label{eq::covariance_dynamics}\\
    &+W_{I,\Delta_1})\Xi_{(I,J)}^T
    +W_{R_{(I,J)}}]A_{J,\Delta_2}^T + W_{J,\Delta_2}\nonumber
\end{align}
where the saltation matrix is evaluated at $\Xi_{(I,J)}=\Xi_{(I,J)}(\bar{t}_{i},A_{I,\Delta_1}\hat{x}(k))$.

A naive approach to updating the covariance through a hybrid transition is to simply use the Jacobian of the reset function instead of the saltation matrix in Eq. \eqref{eq::covariance_dynamics}.
To illustrate the difference between this naive approach and the proposed, we compare using the Jacobian of the reset map instead of the saltation matrix in all experiments. 

\subsection{Hybrid transition during \emph{a posteriori} update}
\label{section:aposteriori}
Next, we consider the case where the measurement update pulls the mean estimate into a guard set \eqref{eq:fullguard}, i.e.\ $\hat{x}(k+1|k+1)\in G_{(I,J)}$ for some $J$.  In that case, the \emph{a posteriori} update is modified by applying the reset to the mean and the saltation update to the covariance after applying the standard update \eqref{eq::kalmangain}--\eqref{eq::postcov},
\begin{align}
    \tilde{x}(k+1|k) &= R_{(I,J)}\hat{x}(k+1|k) \label{eq::meanposterioritransition}\\
    \tilde{\Sigma}(k+1|k) &= \Xi_{(I,J)}\hat{\Sigma}(k+1|k)\Xi_{(I,J)}^T +W_{R_{(I,J)}}
    \label{eq::covposterioritransition}
\end{align}
where the saltation matrix is evaluated at $\Xi_{(I,J)}=\Xi_{(I,J)}(\bar{t}_{i},\hat{x}(k+1|k))$. These $\tilde{x}(k+1|k)$ and $\tilde{\Sigma}(k+1|k)$ are the updated \emph{a posteriori} mean and covariance in the new hybrid domain, $J$. Note that this update is identical to  \eqref{eq::kfresetdynamicupdate}--\eqref{eq::kfcovdynamicupdate}.



\subsection{Extended Kalman Filter}
\label{section:ekf}
Similar to the Kalman filter, the standard Extended Kalman Filter (EKF) \cite[Eqn.~2.1--2.2]{welch1995introduction} can be directly applied for nonlinear hybrid systems when no transition occurs. The nonlinear stochastic dynamics are given by
\begin{align}\allowdisplaybreaks
    x(k+1) &= f_{I,\Delta}(x(k),u(k),\omega(k))\label{eq:nonlineardiscretedyn}\\
    \hat{A}_{I,\Delta} &= D_x f_{I,\Delta}(x(k),u(k),\omega(k))\\
    \hat{W}_{I,\Delta} &= D_\omega f_{I,\Delta}(x(k),u(k),\omega(k))\\
    y(k) &= h_I(x(k),v_I(k)) \\
    \hat{C}_I &= D_x h_I(x(k))
\end{align}
where $f_{I,\Delta}$ is the discrete nonlinear update for the continuous dynamics $F_I$, $\hat{A}_{I,\Delta}$ is the linear approximation of the dynamics, $\hat{W}_{I,\Delta}$ is the linear approximation of the process noise, $h_I$ is the measurement function and $\hat{C}_I$ is the linear approximation of the measurement function.

When there is a hybrid transition during the \emph{a priori}  update, the dynamic updates for the nonlinear transition case are substituted in the same manner as the linear case into \eqref{eq::kf_mean_apriori}--\eqref{eq::covariance_dynamics}. 
When there is a hybrid transition during the \emph{a posteriori} update, the mean update equation \eqref{eq::meanposterioritransition} is applied with the full nonlinear reset map, while the covariance update \eqref{eq::covposterioritransition} is the same for both the linear and nonlinear hybrid systems because the saltation matrix is already a linearization. With these updates, the nonlinear extension to the Salted Kalman Filter follows naturally. 

\subsection{Summary and psuedocode}

The Salted Kalman Filter (SKF) as presented above is summarized in Algorithm~\ref{alg:skf}. Note that the only difference from the standard Kalman Filter algorithm is applying the proposed moment updates when the estimated state satisfies the guard condition (lines 7--11 and 16--20).
The SKF is in many ways similar to the EKF because the saltation matrix is a linearization about the hybrid transition -- if the transition is linear or the prediction is close to the actual then the filter performs well. 
This property holds for the nonlinear Extended SKF as well, and in general this filter suffers from the same pitfalls as the EKF. 
Furthermore, like the EKF this linearization means that the optimal belief may not remain Gaussian, and thus that the filter may fail to have the optimally properties we obtain in the linear case. 

For the measurement update, if a hybrid transition is triggered, the approach presented here simply transforms the already updated estimates. 
However, a more accurate approach might include breaking up the measurement update into sub-updates over each domain. 
In this work, we assume the updates are small enough such that this isn't an issue, but as the measurement update magnitude increases, this may be worth investigating. 
While the extended version of this filter is not optimal, like the EKF, we expect that it will perform well when the covariances and timesteps are relatively small so that the local linearizations hold. Therefore, we expect the performance of the filter to falter when the estimation heavily deviates from the actual trajectory in cases such as initializing the filter far away from the actual starting state, initializing in the wrong mode, or trajectories with grazing impact (when the dynamics are not transverse to the guard).

\begin{algorithm}[t]
\caption{Salted Kalman Filter (SKF)}\label{alg:skf}
\begin{algorithmic}[1]
\State \textbf{input} ($t_{k}$, $x_{k}$, $\Sigma_{k}$, $m_{k}$, $y_{k+1}$)
\State $\hat{t} \gets t_{k}$, $\hat{x} \gets x_{k}$, $\hat{\Sigma} \gets \Sigma_{k}$, $I \gets m_{k}$
\While{$(\hat{t}< t_k+ \Delta)$}
    \State $(\hat{t}^+,\hat{x}) \gets$ integrate $F_I(\hat{t},\hat{x})$ 
    
    \qquad until $(\hat{t}^+ = t_k + \Delta)$ or $(\exists J$ s.t. $\hat{x} \in G_{(I,J)})$ 
    \State $\Delta_1 \gets \hat{t}^+-\hat{t}$, $\hat{t} \gets \hat{t}^+$
      \State $\Sigma_{k} \gets A_{I,\Delta_1}\hat{\Sigma}A^T_{I,\Delta_1}+W_{I,\Delta_1}$  \Comment{\eqref{eq:apriorisigma}}
      \If{$\exists J$ s.t. $\hat{x} \in G_{(I,J)}$} 
        \State $\hat{x} \gets R_{(I,J)}(\hat{t},\hat{x})$ \Comment{\eqref{eq::kfresetdynamicupdate}}
        \State $\hat{\Sigma} \gets \Xi_{(I,J)}\hat{\Sigma}\,\Xi_{(I,J)}^T+W_{R_{(I,J)}}$\Comment{\eqref{eq::kfcovdynamicupdate}}
        \State $I \gets J$ 
\EndIf
\EndWhile\label{simulatewhile}
\State $K \gets \hat{\Sigma}C_{I}^T\left[C_I\hat{\Sigma}C_I^T + V_I\right]^{-1}$ \Comment{\eqref{eq::kalmangain}}
\State $\hat{x} \gets \hat{x} + K\left[y_{k+1} - C_I \hat{x}\right]$ \Comment{\eqref{eq::postmean}}
\State $\hat{\Sigma} \gets \hat{\Sigma} - K C_I\hat{\Sigma}$ \Comment{\eqref{eq::postcov}}
\If{$\exists J$ s.t. $\hat{x} \in G_{(I,J)}$} 
        \State $\hat{x} \gets R_{(I,J)}(\hat{t},\hat{x})$ \Comment{\eqref{eq::meanposterioritransition}}
        \State $\hat{\Sigma} \gets \Xi_{(I,J)}\hat{\Sigma}\,\Xi_{(I,J)}^T +W_{R_{(I,J)}}$ \Comment{\eqref{eq::covposterioritransition}}
        \State  $I \gets J$
\EndIf
\State $t_{k+1} \gets \hat{t}$, $x_{k+1} \gets \hat{x}$, $\Sigma_{k+1} \gets \hat{\Sigma}$,  $m_{k+1} \gets I$
\State \textbf{return} ($t_{k+1}$, $x_{k+1}$, $\Sigma_{k+1}$, $m_{k+1}$)
\end{algorithmic}
\end{algorithm}
\section{Experiments}
\label{section:Experiments}

This section lays out the experimental design (Sec.~\ref{sec:expdesign}) and example systems (Sec.~\ref{sec:hsdef}) that are used to test the utility of the Salted Kalman Filter.

\subsection{Experimental Design}
\label{sec:expdesign}
In the experiments, three different estimation techniques are used: 1) the proposed Salted Kalman Filtering (SKF) algorithm using the saltation matrix to map covariance, 2) the naive mapping using the Jacobian of the reset map (which we call the Jacobian of the Reset Kalman Filter, JKRF, and which follows Algorithm~\ref{alg:skf} but with the saltation matrix $\Xi$ replaced by the Jacobian of the reset map $D_xR$), and 3) a hybrid system Particle Filter (PF), following \cite{koutsoukos2002monitoring}. 
Experiments are performed in simulation to ensure consistency and accurate model knowledge. 
These experiments evaluate the SKF by comparing the mean squared error of the 3 filters in a series of Monte Carlo tests.


For the simulation, the stochastic difference equation, \eqref{eq:nonlineardiscretedyn}, is calculated for each timestep using MATLAB's \texttt{ode45} \cite{shampine2003solving} where the integration follows \eqref{eq:noiseintegration}. 
Ode45 is used to account for the guard zero crossing detection using the MATLAB event location feature.

Tests comparing the Kalman Filters were run with a range of measurement noise, process noise, and time steps.
Tests comparing to the particle filter were run with a range of time steps with a single representative process and measurement noise.
For simplicity the starting covariance, starting mean, reset covariance, chosen measurements, and simulation time were held constant between trials.

The effectiveness of the filter for each trial is evaluated by calculating the mean squared error (MSE) along a simulated trajectory,
\begin{equation}
    \text{MSE} = \frac{1}{K}\sum_{k=1}^{K}\left((x(t_k)-\hat{x}(t_k))^T(x(t_k)-\hat{x}(t_k)\right)
\end{equation}
where $K$ is the number of time steps, $\hat{x}(t_k)$ is the state estimate at time $t_k$, and $x(t_k)$ is the true state at time $t_k$. 
For each measurement noise, process noise, and time step combination, the filter is run on 1000 randomly sampled starting conditions with randomly sampled process noise and randomly sampled measurements.
The same random trials are then passed to each filter for comparison.
Each set of trials are compared using the \emph{sign test} \cite{dixon1946statistical}.
The null hypothesis is that the median difference between the pairs is zero,
\begin{equation}
    H_0: MSE_1 - MSE_2 = 0
    \label{eq::nullhypothesis}
\end{equation}
The sign test is chosen because the data are not normally distributed, which rules out the paired t-test, and are not necessarily symmetric, which rules out the Wilcoxon Signed Rank test.

\subsection{Hybrid System Definitions}
\label{sec:hsdef}
We present experiments for two different hybrid systems: 1) a simpler system which retains a Guassian distribution, Sec.~\ref{sec:constantflow}, and 2) a more complex system with nonlinear non-identity reset maps, nonlinlear dynamics, and a higher dimensional state space, Sec.~\ref{sec:aslip}. 

\subsubsection{Constant Flow}
\label{sec:constantflow}
The simplest hybrid system we examine is the case where there are two hybrid modes that are linearly separated and which have constant, but distinct, dynamics in each mode. The dynamics in the hybrid modes are defined:
\begin{align}
    F_1 &= [1,-1]^T, \quad
    F_2 = [1,1]^T
\end{align}
The guard sets are defined at $x_1=0$, such that the domain of $F_1$ is the left half plane and the domain of $F_2$ is the right half plane (Fig.~\ref{fig:covariance_mappings}).  The reset is an identity map. 
The measurements for this system were chosen to be both states, i.e.,
\begin{equation}
    h_I(x) = \begin{bmatrix}
    1&0\\
    0&1
    \end{bmatrix}x = Cx
\end{equation}

\subsubsection{Asymmetric Spring Loaded Inverted Pendulum (ASLIP)}
\label{sec:aslip}
\begin{figure}
    \centering
    \includegraphics[width =0.9\columnwidth]{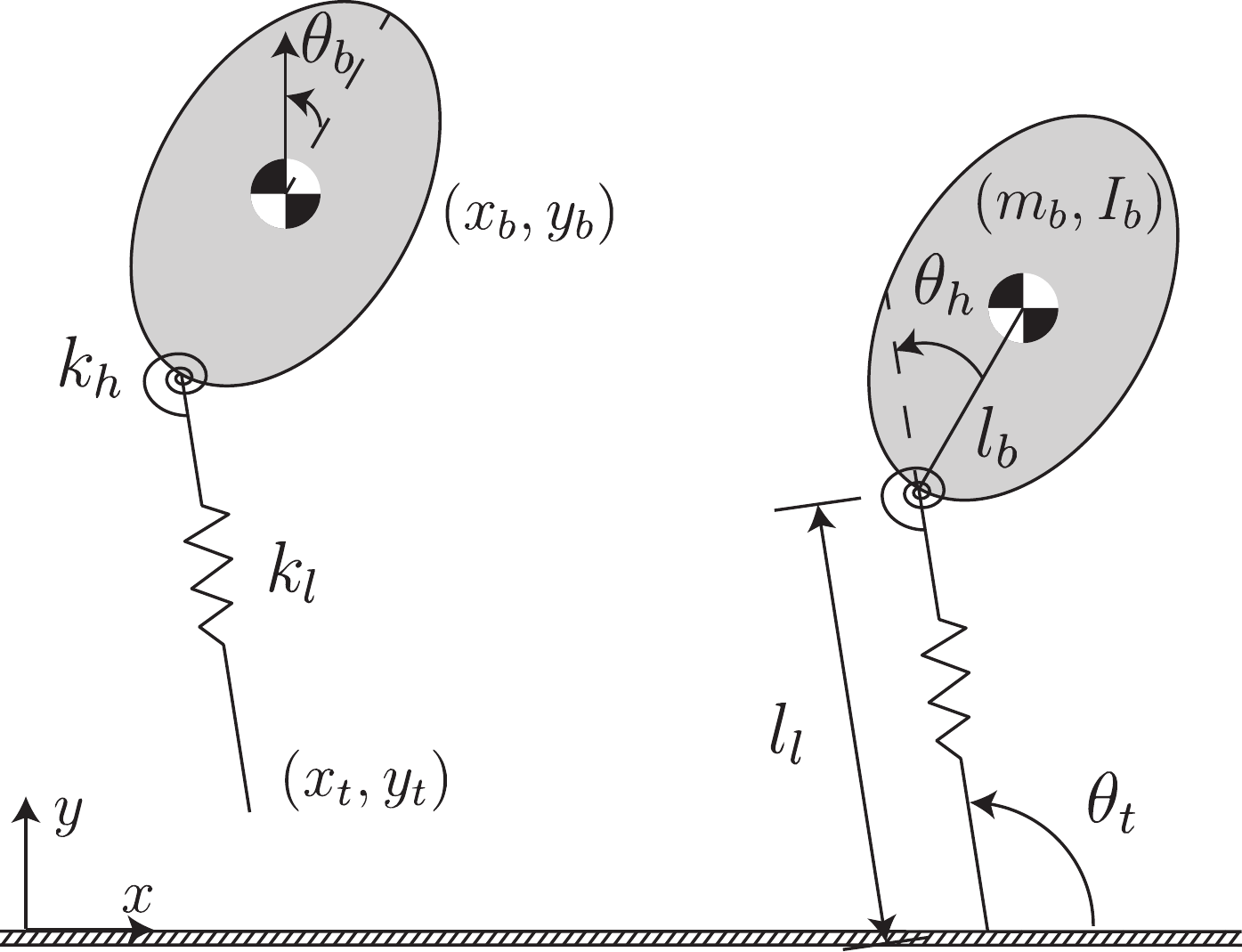}
    \caption{Asymmetric Spring Loaded Inverted Pendulum (ASLIP) diagram showing the aerial phase hybrid mode on the left and the stance phase hybrid mode on the right and their corresponding configuration variables.}
    \label{fig:slipdiagram}
\end{figure}

The asymmetric spring loaded inverted pendulum (ASLIP) system consists of a spring leg, torsional spring hip, and a body with inertia in the sagittal plane as shown in Fig. \ref{fig:slipdiagram}. 
This system is similar to the one in \cite{poulakakis2009slip} and a full derivation for the system dynamics can be found in Appendix~\ref{app:ASLIP}.
This hybrid system is especially useful to analyze because it includes both nonlinear dynamics and non-identity resets.

In this system, the body configuration space is defined to be the position and  orientation of the body $q_b := [x_b,y_b,\theta_b]^T\in \mathbb{R}\times\mathbb{R}\times\mathbb{S}^1$. 
The leg configuration space is defined to be the angle between the toe and the ground, the angle of the hip, and the length of the leg $q_l := [\theta_t,\theta_h,l_l]^T\in \mathbb{S}^1\times\mathbb{S}^1\times\mathbb{R}$, where impact location of the toe defines a pin joint for the body to pivot around.
Once the location of the toe, $q_t = [x_t,y_t]^T\in \mathbb{R}\times\mathbb{R}$, is fixed to a ground location, either configuration can be used to define the full configuration space of the system. 
When the toe position is known, the transformation from the leg configuration to the body configuration is defined as $T_{lb}:(q_l, q_t)\mapsto q_b$, while the inverse mapping is defined as $T_{bl}:(q_b, q_t)\mapsto q_l$. 

Hybrid mode 1 is defined to be when the toe is not in contact with the ground. 
The resulting domain $\mathcal{D}_1$ is chosen to be parameterized by the body's configuration, toe position, and body's velocity.
\begin{equation}
    [x_b,y_b,\theta_b,x_t,y_t,\dot{x}_b,\dot{y}_b,\dot{\theta}_b]^T \in \mathcal{D}_1
\end{equation}
When the toe is in contact with the ground, the hybrid mode is 2. 
The domain $\mathcal{D}_2$ is chosen to be parameterized by the toe angle with the ground, hip angle, the leg extension, toe position, the time derivative of the toe angle, hip angle, and leg extension.
\begin{equation}
    [\theta_t,\theta_h,l_l,x_t,y_t,\dot{\theta}_t,\dot{\theta}_h,\dot{l}_l]^T \in \mathcal{D}_2
\end{equation}
Note that the toe position is augmenting the state rather than being treated as an external parameter because variations in the toe placement affect the other configuration states. 
Because of this, the toe dynamics are constrained relative to the body in domain 1 and relative to the ground contact in domain 2.
These dynamics $F_1, F_2$ are derived using a Lagrangian approach (see Appendix~\ref{app:ASLIP}). 
The system parameters and their experimental values are body mass $m_b = 1$, body inertia $I_b = 1$, leg spring constant $k_l = 1000$, hip spring constant $k_\theta=400$, body length $l_b = 0.5$, acceleration due to gravity $a_g = 9.8$, resting leg length $l_{l0} = 1$, and resting angle of the hip spring $\theta_{h0} = -\frac{\pi}{8}$.

The guard for mode 1 is defined to be when the toe touches the ground, $g_{(1,2)}(t,q,\dot{q}) = y_t$, and the guard for mode 2 is defined to be when the normal force of the toe reaches zero, i.e when the leg spring reaches the resting length, $g_{(2,1)}(t,q,\dot{q}) = l_l - l_{l0}$. 
The reset maps are defined to be the coordinate changes from the body states to the leg states.
\begin{align}\allowdisplaybreaks
    R_{1,2} &= \begin{bmatrix}
    T_{bl}(q_b)\\
    q_t\\
    D_{q_b}T_{bl}(q_b,q_t)\dot{q}_b\\
    \end{bmatrix}\\
    R_{2,1} &= \begin{bmatrix}
    T_{lb}(q_l)\\
    q_t\\
    D_{q_l}T_{lb}(q_l,q_t)\dot{q}_l\\
    \end{bmatrix}
\end{align}

For this system, only measurements of the body states are given. 
This is more realistic, because it is assumed that the hybrid mode is unknown. 
Therefore, in the aerial phase, hybrid mode 1, the measurement function is simply.
\begin{equation}
    h_1(x) = \begin{bmatrix}
    q_b\\
    \dot{q}_b
    \end{bmatrix}
\end{equation}
However, in the stance phase, hybrid mode 2, the states are the leg states and the toes positions and cannot be directly compared against the body measurements. 
Therefore, the measurement function in hybrid mode 2 is the transformation from leg states to body states
\begin{equation}
    h_2(x) = \begin{bmatrix}
    T_{lb}(q_l)\\
    T_{lb}(q_l,q_t,\dot{q}_l)
    
    \end{bmatrix}
\end{equation}

\section{Results}
\label{section:Results}
In this section we present the results of the experiments detailed in the previous section on the example hybrid systems.

\subsection{Constant Flow}
    \begin{figure}[tb]
    \centering
    \includegraphics[width = \columnwidth,clip]{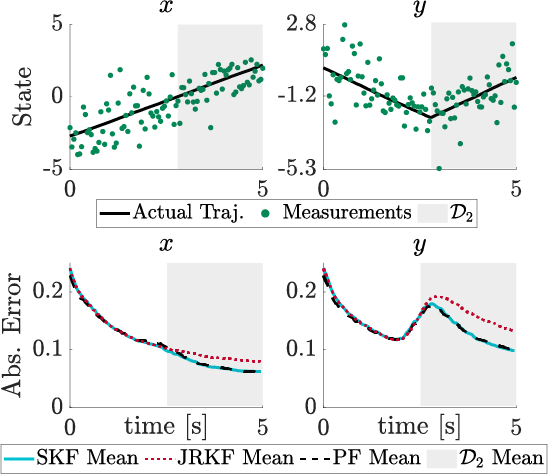}
    \caption{Kalman filter results on the constant flow system. Note that the main differences are just after the transition where the methods differ, but because Kalman Filters are stable these differences disappear as time goes on. Testing conditions for this example are timestep $\Delta = 0.05$s, process noise $\|W_{I,\Delta}\|=0.01\Delta^2$, and measurement noise $\|V_I\| = 1$. Top: For a single trial, the ground truth trajectory (black solid) is shown with the measurements (green dots) and highlighting (gray shaded) when the system is in  $\mathcal{D}_2$. Bottom: Absolute mean error is plotted for the SKF (blue solid), JRKF (red dots), and PF (black dashed) while highlighting (gray shaded) the mean transition time to $\mathcal{D}_2$.}
    \label{fig:kf_constantflow}
\end{figure}
The first experiment uses the constant flow system defined in Sec.~\ref{sec:constantflow} and shown in Fig.~\ref{fig:covariance_mappings}.
The system was simulated for $5$ seconds with 4 different time steps: $\Delta=5$, $1$, $0.1$, and $0.05$ seconds. 
The process covariance levels ranged from $\|W_{I,\Delta}\| = 0.1\Delta^2$ to $0.0001\Delta^2$ and the measurement covariance was swept from $\|V_I\| = 1$ to $0.0001$, both in powers of 10, for a total of $4$ process covariance levels and $5$ measurement covariance levels.
In total, the Monte Carlo simulations for the $80$ parameter sets were tested with $1000$ trials each. An example experiment is shown in Fig.~\ref{fig:kf_constantflow}. Note in particular the difference when comparing the error starting at the hybrid transition.

The result of the Monte Carlo Kalman filter tests were that the SKF performed better than the JRKF for $76$ of the $80$ combinations (to statistical significance $p<0.05$).
In the $4$ remaining cases the filters are statistically indistinguishable, and in none of the experiments did the JRKF outperform the SKF to statistical significance. 
For each of these cases, the time step is large, the measurement noise is low, and the process noise is high, and so both filters depend mostly on the sensors and therefore the difference in dynamic update becomes less important.

For the particle filter experiment, the following parameters were chosen: process noise $\|W_{I,\Delta}\|=0.01\Delta^2$, measurement noise $\|V_I\| = 1$, initial covariance $\Sigma(0) = 0.1I$, and $\Delta=5$, $1$, $0.1$, and $0.05$ seconds.
There was no noise added to reset because the reset map is an identity transformation. 
The particle filter was initialized with between $50$ and $3000$ particles sampled from the initial distribution.

\begin{figure}[tb]
    \centering
    \includegraphics[width = \columnwidth,clip]{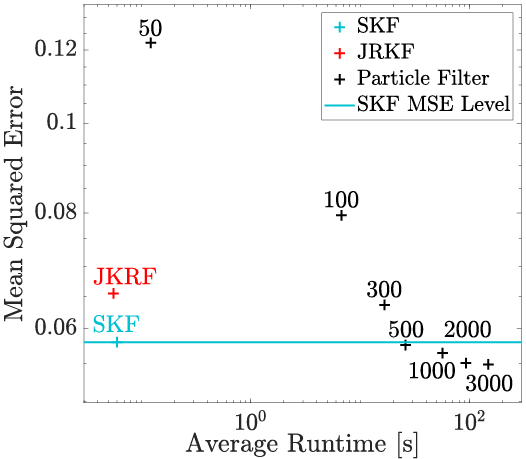}
    \caption{Mean Squared Error versus average runtime for constant flow case with the particle filter ranging from 50 to 3000 particles (black pluses) compared against the JKRF (red plus) and SKF (blue plus, with a constant blue line highlighting the SKF MSE level for comparison). The means were taken from a Monte Carlo Simulation with 1000 trials where $\Delta = 0.05$s, process noise $\|W_{I,\Delta}\|=0.01\Delta^2$, and measurement noise $\|V_I\| = 1$.}
    \label{fig:runtimevsmse}
\end{figure}

The results of the particle filter experiments are shown in Fig.~\ref{fig:runtimevsmse}, where it is clear that the particle filter took significantly higher computation time than the Kalman filters. 
This is expected, because the Kalman filters is only simulating $1$ particle's mean and updating the covariance with matrix computations. 
Starting only at $1000$ particles did the PF perform statistically better than the SKF, with a decrease of $2.7\%$ MSE at the cost of taking $941$ times longer to compute. 
At $2000$ particles, the decrease is $5.2\%$ in MSE and the computation required $1736$ times the SKF's computation time. Increasing the number of particles to $3000$ did not result in a statistically significant improvement over $2000$ and so for further comparison with the SKF and JKRF, the number of particles was held constant at $2000$.


Considering the effect of the time step on the particle filter experiments, at the largest time step ($\Delta = 5$s) the MSE of the SKF and the PF are statistically indistinguishable. 
For the smaller time steps ($\Delta = 1$s, $0.1$s, and $0.05$s), the PF has lower MSE than the SKF $(p<0.05)$. We hypothesize that this is due to the assumption in the SKF that the majority of the probability mass transitions together during a single timestep. The SKF performs comparably worse when the timesteps are small and the distribution is split across a hybrid transition. 
To test this hypothesis, we compare the time step levels to the time it takes for this system to transition $99\%$ of the probability mass at the transition time as shown in Table~\ref{table::transition_table}. We find that if the time to transition $\Delta_T$ was less than the timestep duration $\Delta$, then no increase in performance was observed with the PF.

\begin{table}[tb]
\caption{The covariance magnitude at the time of transition $\|\Sigma(\bar{t}_{i})\|$ and the ratio between the time it takes to transition $99\%$ of the probability mass for each timestep level $\Delta_T$ and the current timestep length $\Delta$ for the constant flow system with process noise $\|W_{I,\Delta}\|=0.1\Delta^2$ and measurement noise $\|V_I\| = 1$. Trials where the PF had a statistically lower MSE than the SKF are marked with a~$^*$.}
\centering
\vspace{.5em}
\begin{tabular}{l c c}
\hline
$\Delta$  & $\|\Sigma(t_i)\|$ & $\Delta_T/\Delta$\\
\hline
$5s$& $0.16$& $0.38$\\
$1s^*$& $0.10$& $1.5$\\
$0.1s^*$& $ 0.028$& $7.9$\\
$0.05s^*$& $0.015$& $12$\\[2pt]
\hline
\label{table::transition_table}
\end{tabular}
\end{table}

\subsection{ASLIP}
    \begin{figure*}[tb]
    \centering
    \includegraphics[width = \textwidth]{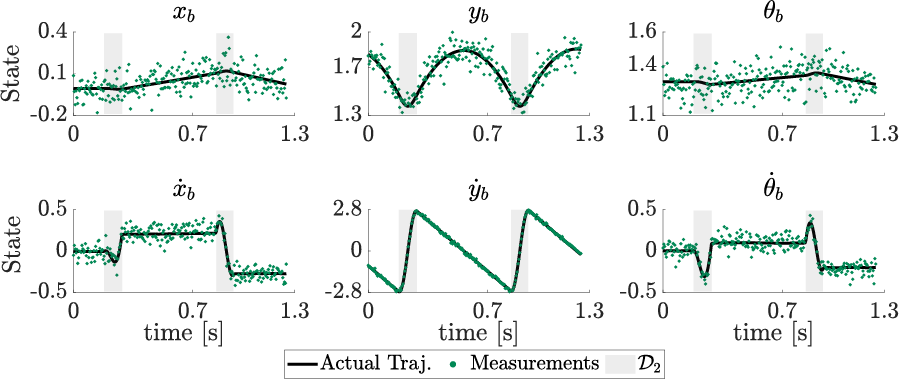}
    \includegraphics[width = \textwidth]{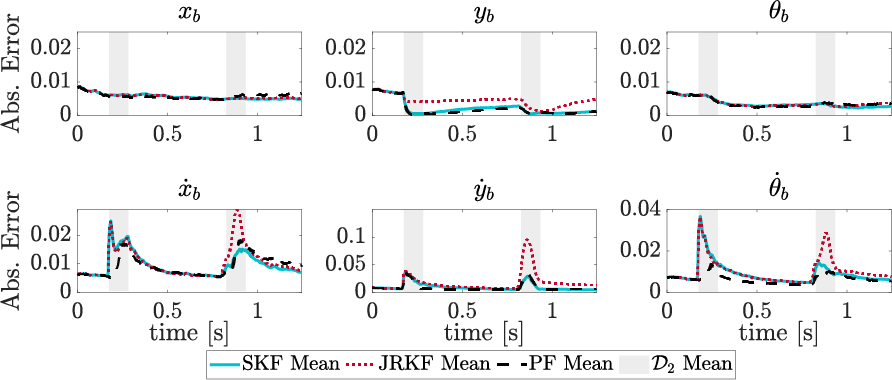}
    \caption{ASLIP Kalman filter results comparing the SKF to the JRKF. Note that the main differences between the methods are at the transitions and also that the improvement is in one direction (here, mostly in the vertical body position $y_b$) because the saltation matrix is different from the Jacobian of the reset map by a rank 1 update. Testing conditions for this example are timestep $\Delta = 0.005$s, measurement noise $\|V_I\| = 0.005$, and process noise $\|W_{I,\Delta}\|=0.01\Delta^2$. Top:  For a single trial, the ground truth trajectory (black solid) is shown with the measurements (green dots) and highlighting (gray shaded) when the system is in $\mathcal{D}_2$. Bottom: Absolute mean error is plotted for the SKF (blue solid), JRKF (red dots), and PF (black dashed) while highlighting (gray shaded) the mean transition times to $\mathcal{D}_2$.}
    \label{fig:kf_slip}
\end{figure*}
The Kalman filtering and particle filtering experiments were also run on the ASLIP system, defined in Sec.~\ref{sec:aslip}.
For these tests, we simulated the dynamics for $1.25$ seconds which resulted in $2$ jumps (4 hybrid transitions).
Experiment time steps were set at $\Delta \in \{ 0.03$, $0.01$, $0.005$, $0.001 \}$ seconds. 
The process noise covariance levels were $\|W_{I,\Delta}\| = 0.01\Delta^2$, $0.001\Delta^2$, and $0.0001\Delta^2$, and the measurement noise covariance levels were $\|V_I\|= 0.005$, $0.001$, and $0.0001$.
The initial covariance was set to be $1\times 10^{-4}I$, where the noise in the toe position was set to match the constraint between the body configuration and the toe (as the toe position is correlated to the body states). 
Reset noise is not applied because there is no uncertainty in the coordinate transformation. 
 
In total, the Monte Carlo simulation for the 36 parameter sets that were tested with 100 trials each.
An example experiment is shown in Fig.~\ref{fig:kf_slip}.
The result of these tests were that the SKF performed better than the JRKF for all 36 combination with statistical significance $(p<0.001)$.
While the SKF performs better than the JRKF on average over all states, this does not indicate that the SKF performs better than the JRKF in all coordinates for each timestep.
In several of the Monte Carlo simulations, the mean absolute error peaked above the JRKF's mean in $\dot{x}_b$, $\dot{y}_b$, or $\dot{\theta}_b$ for several timesteps -- generally after the first touchdown.
However, one consistent difference that was seen in all simulations was that SKF had sustained improvements in the vertical body position $y_b$. 
The difference between the saltation matrix and the Jacobian of the reset map on impact is in the column associated with the vertical height $y_b$.
Therefore, the improvements in $y_b$ are expected because the dynamics along this axis are accounted for.

For the particle filter experiment, $30,000$ particles were used and the following testing parameters were chosen: process noise $\|W_{I,\Delta}\|=0.01\Delta^2$, measurement noise $\|V_I\| = 0.005$, and $\Delta=0.005$ seconds. 
An example run with these parameters are shown in the top plot of Fig. \ref{fig:kf_slip} and the filter performance is shown in the lower plot. As with the constant flow system,  the filters again perform similarly for each state away from hybrid transitions and the differences are magnified near hybrid transitions. 

The result of this experiment was that the SKF has a lower MSE than the particle filter with statistical significance ($p<0.001$) over the 100 trials. We believe that the particle filters performance can be improved to be better than or equal to the performance of the SKF by increasing the number of particles. However, at $30,000$ particles the computation time is already unwieldy, taking on average $5200$ seconds to simulate a 1.25s experiment. Similar to constant flow example, the hybrid particle filter takes significantly longer ($\times22000$) to run than the SKF.

\section{Conclusion}
\label{section:Conclusion}
In this paper, we created a new Kalman filtering algorithm which allows estimation on hybrid dynamical systems with state-defined transitions, including an extended Kalman filter variant which can handle nonlinear dynamics with non-identity reset maps.
This ``Salted Kalman Filter'' was validated on both a linear and nonlinear system and compared against both a particle filter and the ``Jacobian of the reset map'' counterpart.

The results show that using our proposed method is statistically better than or equivalent to the naive method in all tested cases. However, both Kalman filters perform well and have relatively low mean squared error.
We believe this is because the naive solution and our proposed method have the same mean update and algorithm structure, the fact that they both perform well highlights the importance of having an accurate update for the mean as well as handling each transition case in the algorithm. 
When comparing against a hybrid particle filter for the constant flow case, the SKF is statistically indistinguishable when we are able to closely approximate that the probability distribution stays Gaussian and that the majority of the probability mass transitions in a single or several time steps.
When the assumption that the probability mass transitions over a small number of time steps is broken, the particle filter outperformed the SKF, but the largest increase in performance was small ($5.2\%$) especially compared to the 1736 times increase in computation time.

For the more complex ASLIP system, the SKF performed statistically better than the $30,000$ particle filter when comparing MSE. 
However, we believe that with enough particles the particle filter should be better than the SKF, though increasing the number of particles would increase the computation time.

The proposed method, similar to the extended Kalman Filter, suffers when model uncertainty is added to the hybrid dynamical system, when the local approximation is violated, or when the noise is non-Gaussian. 
Overall, like an extended Kalman filter, if the estimate diverges from the actual trajectory (i.e.\ the estimate is initialized far away from the actual, the starting mode is incorrect, or if an incorrect transition is made) the performance of the filter will suffer.
Incorrect mode transitions are mitigated by the class of hybrid dynamical systems that are considered which require transverse guards (Assumption \ref{asm:salt-exists}).
In cases where the non-linearity, non-localness, or non-Gaussianness are significant, a hybrid particle filter or other particle filtering approaches may be more appropriate, but will be accompanied with a respective increase in computation complexity.
For a smooth system, an unscented Kalman filter may be used in place of an extended Kalman filter if the local assumption is not valid. However, using an unscented Kalman filter for a hybrid dynamical system may not transfer well because the sampled sigma points may end up past the guard.

Note that while using the saltation matrix captures the update for the covariance to first order, the saltation matrix is model-dependent, and may require significant effort to obtain in practice in order to use \eqref{eq:saltationmatrix} directly.
However, as the saltation matrix is a linear map relating pre- and post-transition states, regression techniques may be able to approximate it with measured data without the need for complete (and differentiable) models of the hybrid system.

While this is a good start to developing an online hybrid state estimation system, there is still further work needed to improve the estimation. 
Our method does not explicitly reason about the probability of a state or measurement being in a particular hybrid mode or guard, and an extension that reasons about this probability will be covered in future work. 
Additionally, future work is required to cover distributions which pass through intersections of hybrid guards, in which case an extension based on the Bouligand derivative \cite {burden2016event,scholtes2012introduction} could be used to capture the propagation of uncertainty.

\appendix
\section{Derivation of the ASLIP system}
\label{app:ASLIP}
The change of coordinate functions are,
\begin{align}\allowdisplaybreaks
&T_{lb}(q_b,q_t) = \begin{bmatrix}
    l_l \cos(\theta_t)+l_b\cos(\theta_t+\theta_h)+x_t\\
    l_l \sin(\theta_t)+l_b\sin(\theta_t+\theta_h)+y_t\\
    \theta_t+\theta_h
    \end{bmatrix} \\
&T_{bl}(q_{b},q_t) =  \\
&\quad\begin{bmatrix}
    \text{atan}\left(\frac{y_b-(l_b\sin(\theta_b)+y_t)}{x_b -(l_b\cos(\theta_b)+x_t)}\right)\\
    \theta_b - \text{atan}\left(\frac{y_b-(l_b\sin(\theta_b)+y_t)}{x_b -(l_b\cos(\theta_b)+x_t)}\right)\\
    \sqrt{(y_b-l_b\sin(\theta_b)-y_t)^2 + (x_b -l_b\cos(\theta_b)-x_t)^2}
    \end{bmatrix}
    \nonumber 
\end{align}
The differential mappings are defined via chain rule
\begin{align}\allowdisplaybreaks
    \begin{bmatrix}
    \dot{q}_b\\
    \dot{q}_t
    \end{bmatrix} &= D_{q_b, q_t} T_{lb}(q_b,q_t)\begin{bmatrix}
    \dot{q}_l\\
    \dot{q}_t
    \end{bmatrix}\\
    \begin{bmatrix}
    \dot{q}_l\\
    \dot{q}_t
    \end{bmatrix}&= D_{q_l, q_t} T_{bl}(q_l,q_t)\begin{bmatrix}
    \dot{q}_b\\
    \dot{q}_t
    \end{bmatrix}
\end{align}
Since the toe is massless, the velocity of the toe is assumed to be zero when mapping velocities and is therefore removed from the differential mapping.
\begin{align}
    \dot{q}_l &= D_{q_b} T_{lb}(q_b,q_t)\dot{q}_b, \quad
    \dot{q}_b = D_{q_l} T_{bl}(q_l,q_t)\dot{q}_l
\end{align}
The dynamics for mode 1 are ballistic dynamics for the center of mass and because the toe is massless, both the hip and leg springs are kept at their resting locations $\theta_{h0}$ and $l_{l0}$ respectively. Therefore, while in mode 1, the toe is kinematically constrained by the body configuration. Define $T_{bt}:q_b\mapsto q_t$ to be the transformation from the body configuration to the toe configuration
\begin{equation}
    T_{bt}(q_b) = \begin{bmatrix}
     x_b - l_b\cos(\theta) - l_{l0}\cos(\theta_{h0} - \theta_b)\\
    y_b - l_b\sin(\theta) + l_{l0}\sin(\theta_{h0} - \theta_b)
    \end{bmatrix}
\end{equation}
The velocity constraint is enforced through the differential mapping of $T_{bt}$
\begin{equation}
    \dot{q}_t = DT_{bt}(q_b)\dot{q}_b = \!\left[\begin{array}{c}
     \dot{x}_b + \dot{\theta}_b( l_b\sin\theta - l_{l0}\sin(\theta_{h0} - \theta_b))\\
 \dot{y}_b - \dot{\theta}_b( l_b\cos\theta + l_{l0}\cos(\theta_{h0} - \theta_b))\!
    \end{array}\!\right]
\end{equation}
Therefore, the dynamics for mode 1 are
\begin{equation}
    F_1 = \begin{bmatrix}
    \dot{x}_b\\
    \dot{y}_b\\
    \dot{\theta}_b\\
     \dot{x}_b + \dot{\theta}_b( l_b\sin(\theta) - l_{l0}\sin(\theta_{h0} - \theta_b))\\
 \dot{y}_b - \dot{\theta}_b( l_b\cos(\theta) + l_{l0}\cos(\theta_{h0} - \theta_b))\\
    0\\
    -a_g\\
    0
    \end{bmatrix}
\end{equation}
The dynamics for mode 2 are derived using Lagrangian dynamics where the Lagrangian is defined to be the difference between the kinetic and potential energy.
\begin{align}
    \mathcal{L} &= \frac{1}{2}(m_b\dot{x}_b^2 + m_b\dot{y}_b^2 + I_b\dot{\theta}_b^2)\nonumber\\
    &\qquad - m_b a_g(l_l \sin(\theta_t)-l_b\sin(\theta_t+\theta_h)) \nonumber\\
    &\qquad - \frac{1}{2}(k(l_{l0}-l_l)^2 + k_h(\theta_{h0}-\theta_h)^2)
\end{align}
The body states are transformed to the leg states using $T_{bl}$ and $DT_{bl}$. Also, in this example, the toe cannot penetrate the ground and a no slip condition is added. Therefore, the dynamics for the toe are calculated separately.



\bibliographystyle{plain}
\bibliography{references.bib}

\begin{thebibliography}{10}

\bibitem{aizerman1958determination}
Mark~A Aizerman and Felix~R Gantmacher.
\newblock Determination of stability by linear approximation of a periodic
  solution of a system of differential equations with discontinuous right-hand
  sides.
\newblock {\em The Quarterly Journal of Mechanics and Applied Mathematics},
  11(4):385--398, 1958.

\bibitem{Back_Guckenheimer_Myers_1993}
Allen Back, J.~M. Guckenheimer, and Mark Myers.
\newblock A dynamical simulation facility for hybrid systems.
\newblock In {\em Hybrid Systems}, volume 736 of {\em Lecture Notes in Computer
  Science}, pages 255--267. Springer Berlin / Heidelberg, 1993.

\bibitem{balluchi2013design}
Andrea Balluchi, Luca Benvenuti, Maria~D Di~Benedetto, and Alberto
  Sangiovanni-Vincentelli.
\newblock The design of dynamical observers for hybrid systems: Theory and
  application to an automotive control problem.
\newblock {\em Automatica}, 49(4):915--925, 2013.

\bibitem{balluchi2002design}
Andrea Balluchi, Luca Benvenuti, Maria~D Di~Benedetto, and Alberto~L
  Sangiovanni-Vincentelli.
\newblock Design of observers for hybrid systems.
\newblock In {\em International Workshop on Hybrid Systems: Computation and
  Control}, pages 76--89. Springer, 2002.

\bibitem{barhoumi2012observer}
Nabil Barhoumi, Faouzi Msahli, Mohamed Djema{\"\i}, and Krishna Busawon.
\newblock Observer design for some classes of uniformly observable nonlinear
  hybrid systems.
\newblock {\em Nonlinear Analysis: Hybrid Systems}, 6(4):917--929, 2012.

\bibitem{bernardo2008piecewise}
Mario Bernardo, Chris Budd, Alan~Richard Champneys, and Piotr Kowalczyk.
\newblock {\em Piecewise-smooth dynamical systems: theory and applications},
  volume 163.
\newblock Springer Science \& Business Media, 2008.

\bibitem{biggio2014accurate}
Matteo Biggio, Federico Bizzarri, Angelo Brambilla, and Marco Storace.
\newblock Accurate and efficient psd computation in mixed-signal circuits: A
  time-domain approach.
\newblock {\em IEEE Transactions on Circuits and Systems II: Express Briefs},
  61(11):905--909, 2014.

\bibitem{mitcheetahstateestimation2018}
G.~{Bledt}, P.~M. {Wensing}, S.~{Ingersoll}, and S.~{Kim}.
\newblock Contact model fusion for event-based locomotion in unstructured
  terrains.
\newblock In {\em 2018 IEEE International Conference on Robotics and Automation
  (ICRA)}, pages 4399--4406, 2018.

\bibitem{bloesch2013state}
Michael Bloesch, Marco Hutter, Mark~A Hoepflinger, Stefan Leutenegger,
  Christian Gehring, C~David Remy, and Roland Siegwart.
\newblock State estimation for legged robots-consistent fusion of leg
  kinematics and {IMU}.
\newblock In {\em Robotics: Science and Systems}, pages 17--24, 2012.

\bibitem{blom1988interacting}
Henk~AP Blom and Yaakov Bar-Shalom.
\newblock The interacting multiple model algorithm for systems with markovian
  switching coefficients.
\newblock {\em IEEE Transactions on Automatic Control}, 33(8):780--783, 1988.

\bibitem{blom2004particle}
Henk~AP Blom and Edwin~A Bloem.
\newblock Particle filtering for stochastic hybrid systems.
\newblock In {\em IEEE Conference on Decision and Control}, volume~3, pages
  3221--3226, 2004.

\bibitem{burden2018contraction}
Samuel~A. Burden, Thomas Libby, and Samuel~D. Coogan.
\newblock On contraction analysis for hybrid systems, 2018.
\newblock arXiv:1811.03956.

\bibitem{burden2016event}
Samuel~A Burden, S~Shankar Sastry, Daniel~E Koditschek, and Shai Revzen.
\newblock Event--selected vector field discontinuities yield
  piecewise--differentiable flows.
\newblock {\em SIAM Journal on Applied Dynamical Systems}, 15(2):1227--1267,
  2016.

\bibitem{dixon1946statistical}
Wilfrid~J Dixon and Alexander~M Mood.
\newblock The statistical sign test.
\newblock {\em Journal of the American Statistical Association},
  41(236):557--566, 1946.

\bibitem{eras2019equality}
Wendy~Y Eras-Herrera, Alexandre~R Mesquita, and Bruno~OS Teixeira.
\newblock Equality-constrained state estimation for hybrid systems.
\newblock {\em IET Control Theory \& Applications}, 13(13):2018--2028, 2019.

\bibitem{ferrari2002moving}
Giancarlo Ferrari-Trecate, Domenico Mignone, and Manfred Morari.
\newblock Moving horizon estimation for hybrid systems.
\newblock {\em IEEE Transactions on Automatic Control}, 47(10):1663--1676,
  2002.

\bibitem{goebel2009hybrid}
Rafal Goebel, Ricardo~G Sanfelice, and Andrew~R Teel.
\newblock Hybrid dynamical systems.
\newblock {\em IEEE control systems magazine}, 29(2):28--93, 2009.

\bibitem{hartley2020contact}
Ross Hartley, Maani Ghaffari, Ryan~M Eustice, and Jessy~W Grizzle.
\newblock Contact-aided invariant extended kalman filtering for robot state
  estimation.
\newblock {\em The International Journal of Robotics Research}, 39(4):402--430,
  2020.

\bibitem{hirsch2012differential}
Morris~W Hirsch, Stephen Smale, and Robert~L Devaney.
\newblock {\em Differential equations, dynamical systems, and an introduction
  to chaos}.
\newblock Academic press, 2012.

\bibitem{hiskens2000trajectory}
Ian~A Hiskens and MA~Pai.
\newblock Trajectory sensitivity analysis of hybrid systems.
\newblock {\em IEEE Transactions on Circuits and Systems I: Fundamental Theory
  and Applications}, 47(2):204--220, 2000.

\bibitem{hwang2006state}
Inseok Hwang, Hamsa Balakrishnan, and Claire Tomlin.
\newblock State estimation for hybrid systems: applications to aircraft
  tracking.
\newblock {\em IET Proceedings-Control Theory and Applications},
  153(5):556--566, 2006.

\bibitem{jeffrey2014dynamics}
Mike~R Jeffrey.
\newblock Dynamics at a switching intersection: Hierarchy, isonomy, and
  multiple sliding.
\newblock {\em SIAM Journal on Applied Dynamical Systems}, 13(3):1082--1105,
  2014.

\bibitem{johnson2016hybrid}
Aaron~M Johnson, Samuel~A Burden, and Daniel~E Koditschek.
\newblock A hybrid systems model for simple manipulation and self-manipulation
  systems.
\newblock {\em The International Journal of Robotics Research},
  35(11):1354--1392, 2016.

\bibitem{Joyce2012}
D.~Joyce.
\newblock {On manifolds with corners}.
\newblock In {\em Advances in Geometric Analysis}, volume~21 of {\em Advanced
  Lectures in Mathematics}, pages 225--258. International Press of Boston,
  Inc., 2012.

\bibitem{khalil2002nonlinear}
Hassan~K Khalil and Jessy~W Grizzle.
\newblock {\em Nonlinear systems}, volume~3.
\newblock Prentice hall Upper Saddle River, NJ, 2002.

\bibitem{koutsoukos2002monitoring}
Xenofon Koutsoukos, James Kurien, and Feng Zhao.
\newblock Monitoring and diagnosis of hybrid systems using particle filtering
  methods.
\newblock In {\em International Symposium on Mathematical Theory of Networks
  and Systems}, 2002.

\bibitem{koval2017manifold}
Michael~C. Koval, Nancy~S. Pollard, and Siddhartha~S. Srinivasa.
\newblock Pose estimation for planar contact manipulation with manifold
  particle filters.
\newblock {\em The International Journal of Robotics Research}, 34(7):922--945,
  2015.

\bibitem{Lee2012}
J.~M. Lee.
\newblock {\em Introduction to smooth manifolds}.
\newblock Springer--Verlag, New York, 2012.

\bibitem{leine2013dynamics}
Remco~I Leine and Henk Nijmeijer.
\newblock {\em Dynamics and bifurcations of non-smooth mechanical systems},
  volume~18.
\newblock Springer Science \& Business Media, 2013.

\bibitem{LygerosJohansson2003}
J.~Lygeros, K.~H. Johansson, S.~N. Simic, J.~Zhang, and S.~S. Sastry.
\newblock {Dynamical properties of hybrid automata}.
\newblock {\em IEEE Transactions on Automatic Control}, 48(1):2--17, 2003.

\bibitem{poulakakis2009slip}
I.~{Poulakakis} and J.~W. {Grizzle}.
\newblock The spring loaded inverted pendulum as the hybrid zero dynamics of an
  asymmetric hopper.
\newblock {\em IEEE Transactions on Automatic Control}, 54(8):1779--1793, 2009.

\bibitem{scholtes2012introduction}
Stefan Scholtes.
\newblock {\em Introduction to piecewise differentiable equations}.
\newblock Springer Science \& Business Media, 2012.

\bibitem{shampine2003solving}
Lawrence~F Shampine, Ian Gladwell, Larry Shampine, and S~Thompson.
\newblock {\em Solving {ODEs} with {MATLAB}}.
\newblock Cambridge university press, 2003.

\bibitem{simic2000towards}
Slobodan~N Simi{\'c}, Karl~Henrik Johansson, Shankar Sastry, and John Lygeros.
\newblock Towards a geometric theory of hybrid systems.
\newblock In {\em International Workshop on Hybrid Systems: Computation and
  Control}, pages 421--436. Springer, 2000.

\bibitem{skaff2005contextbased}
Sarjoun Skaff, Alfred Rizzi, Howie Choset, and Pei-Chun Lin.
\newblock A context-based state estimation technique for hybrid systems.
\newblock In {\em IEEE International Conference on Robotics and Automation},
  pages 3935--3940, April 2005.

\bibitem{thrun2002probabilistic}
Sebastian Thrun.
\newblock Probabilistic robotics.
\newblock {\em Communications of the ACM}, 45(3):52--57, 2002.

\bibitem{varin2018constrained}
Patrick Varin and Scott Kuindersma.
\newblock A constrained kalman filter for rigid body systems with frictional
  contact.
\newblock In {\em International Workshop on the Algorithmic Foundations of
  Robotics (WAFR)}, 2018.

\bibitem{welch1995introduction}
Greg Welch and Gary Bishop.
\newblock An introduction to the kalman filter.
\newblock Technical Report 95--041, Department of Computer Science, University
  of North Carolina at Chapel Hill, 1995.
\newblock {U}pdated: July 24, 2006.

\bibitem{ZHANG2020}
Jize Zhang, Andrew~M. Pace, Samuel~A. Burden, and Aleksandr Aravkin.
\newblock Offline state estimation for hybrid systems via nonsmooth variable
  projection.
\newblock {\em Automatica}, 115:108871, 2020.

\end{thebibliography}
\end{document}